\pdfoutput=1

\documentclass[11pt]{article}

\usepackage[final]{acl}

\usepackage{times}
\usepackage{latexsym}
\usepackage{amsmath}
\usepackage{tcolorbox}

\usepackage[T1]{fontenc}

\usepackage[utf8]{inputenc}

\usepackage{microtype}

\usepackage{inconsolata}

\usepackage{graphicx}

\title{\textit{The `aftermath' of compounds}: Investigating Compounds and their Semantic Representations}

\author{Swarang Joshi \\
    International Institute of Information Technology, Hyderabad, India\\
  \texttt{swarang.joshi@research.iiit.ac.in} \\
}

\begin{document}
\maketitle



\begin{abstract}
This study investigates how well computational embeddings align with human semantic judgments in the processing of English compound words. We compare static word vectors (GloVe) and contextualized embeddings (BERT) against human ratings of lexeme meaning dominance (LMD) and semantic transparency (ST) drawn from a psycholinguistic dataset. Using measures of association strength (Edinburgh Associative Thesaurus), frequency (BNC), and predictability (LaDEC), we compute embedding-derived LMD and ST metrics and assess their relationships with human judgments via Spearman’s correlation and regression analyses. Our study confirms that contextualized embeddings (BERT) 
better mirror human semantic transparency judgments than 
static embeddings (GloVe)\footnote{All code, data, models, and detailed hyperparameters configurations will be publicly released for reproducibility.}. Specifically, BERT's ST values 
show stronger correlation with human annotations (r=0.23 
for frequency, r=0.10 for predictability) and ST predictions 
that more closely align with the expected range (BERT: 3.31-4.25 
vs. human: 4.04-4.93), compared to GloVe's compressed range 
(1.62-3.16). BERT's LMD values
also approximate the human midpoint (5.0) more closely than 
GloVe's representations. And that predictability ratings are strong predictors of semantic transparency in both human and model data. These findings advance computational psycholinguistics by clarifying the factors that drive compound word processing and offering insights into embedding-based semantic modeling.
\end{abstract}


\section{Introduction}
Compound words, such as \emph{teacup} or \emph{bluebird}, pose a unique challenge for both psycholinguistic theory and computational semantics. They consist of two or more free morphemes whose combined meaning may be transparent, as in \emph{teacup}, or less predictable, as in \emph{butterfly}. Psycholinguistic research has long investigated how human readers decompose and interpret compounds, focusing on measures like lexeme meaning dominance (LMD) and semantic transparency (ST) to quantify how strongly constituents contribute to overall meaning \cite{Juhasz2015}. LMD quantifies which constituent 
(left or right) contributes more strongly to the compound's 
overall meaning, rated on a 1-9 scale where values <5 indicate 
left-constituent dominance, 5 represents equal contribution, 
and >5 indicates right-constituent dominance. 
ST measures how readily the compound's 
meaning can be inferred from its constituents, rated on a 1-7 
scale where higher values indicate greater transparency.

With the advent of word embeddings, researchers have begun to probe whether static and contextualized vector representations capture such human semantic intuitions. \citet{buijtelaar-pezzelle-2023-psycholinguistic} pioneered an analysis using BERT embeddings, demonstrating that contextual models may better reflect psycholinguistic patterns than static models like GloVe. However, questions remain about which linguistic factors—frequency, predictability, and associative strength—most robustly predict human judgments and model-derived metrics across embedding types.

In this paper, we extend prior work by systematically comparing GloVe and BERT representations on a shared psycholinguistic dataset of 628 compounds annotated for LMD and ST. We integrate factor ratings from established resources—the Edinburgh Associative Thesaurus\cite{ea-thesaurus}, the Large Database of English Compounds (LaDEC) \cite{Gagné2019}, and the British National Corpus (BNC)—and conduct correlation and regression analyses to evaluate the relative contributions of association, frequency, and predictability. Our contributions are threefold:
\begin{enumerate}
    \item We provide a comprehensive comparison of static versus contextual embeddings in modeling human compound processing.
    \item We identify which linguistic factors most strongly drive embedding-based LMD and ST metrics and their alignment with human data.
    \item We offer recommendations for embedding selection and feature integration in computational psycholinguistics.
\end{enumerate}

\section{Methodology}
We used pre-trained versions of GloVe and BERT to obtain word embeddings.
The Edinburgh Associative Thesaurus \cite{ea-thesaurus} and LaDEC: Large database of English compounds\cite{Gagné2019} were used to get values of the factors - association strength, frequency, and predictability rating.

\subsection{Embedding Extraction}

We used the 300-dimensional GloVe vectors trained on 
840B tokens. Each compound and constituent was extracted as 
its static vector representation. We used bert-base-uncased \cite{devlin2019bertpretrainingdeepbidirectional} (12 layers, 768 dimensions) from Transformers \cite{wolf2020huggingfacestransformersstateoftheartnatural}.

Contextualized and non-contextualized representations of compounds
and their constituent lexemes were obtained.  Cosine similarities between compounds and their constituent lexemes to model lexeme meaning dominance (LMD) and semantic transparency (ST) were computed using the formulae mentioned in \cite{buijtelaar-pezzelle-2023-psycholinguistic}, and MAE and Spearman’s correlation against human-annotated values were evaluated.

Following \citet{buijtelaar-pezzelle-2023-psycholinguistic}, we computed LMD and ST using:

\begin{tcolorbox}[colback=gray!5, colframe=black!75, boxrule=0.5pt, arc=2mm]
\begin{align*}
\text{LMD} &= \left| \cos(\mathbf{v}_c, \mathbf{v}_l) - \cos(\mathbf{v}_c, \mathbf{v}_r) \right| \times 4 + 5\\
\text{ST} &= \frac{\cos(\mathbf{v}_c, \mathbf{v}_l) + \cos(\mathbf{v}_c, \mathbf{v}_r)}{2} \times 3.5
\end{align*}
\end{tcolorbox}

\noindent where $\cos(\mathbf{v}_a, \mathbf{v}_b)$ computes cosine similarity between vectors $\mathbf{v}_a$ and $\mathbf{v}_b$, with subscripts $c$, $l$, $r$ denoting compound, left constituent, and right constituent embeddings.




\subsection{Metrics}
Spearman’s correlation between the factors and the LMD and ST values for human annotations, Glove and BERT embeddings was calculated and regressors were trained to predict the LMD and ST values using those factors. While the association strength and frequency of only the compound was considered, the predictability rating for the lexemes were also considered. 

We use Spearman's correlation to measure the strength and direction of the monotonic relationship between individual linguistic factors (association, frequency, and predictability) and our dependent variables (LMD and ST). This allows us to identify which factors have a significant standalone association with human and model-derived semantic judgments.

We then use regression analysis to assess the predictive power of these factors. The R² score from the regressors reveals the proportion of variance in LMD and ST that can be explained by a single factor, offering deeper insight into its explanatory and predictive utility beyond simple association.


\begin{figure}[t]
    \centering
    \includegraphics[width=0.48\textwidth]{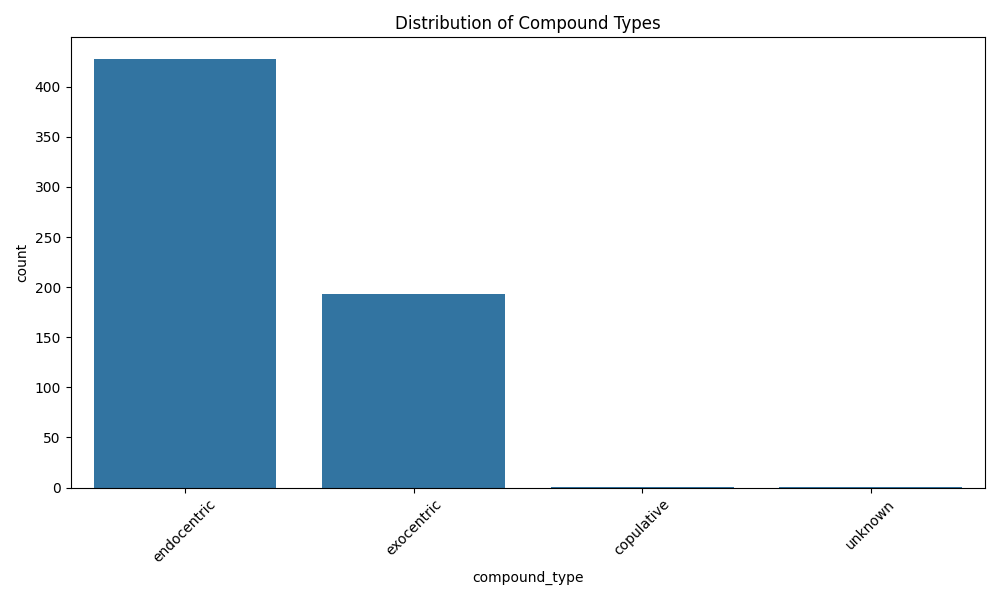}
    \caption{Compound type distribution in dataset (n=628): 68\% endocentric, 31\% exocentric, <1\% copulative.}
    \label{fig:ctypedistro}
\end{figure}

\begin{figure*}
    \centering
    \includegraphics[width=1\textwidth]{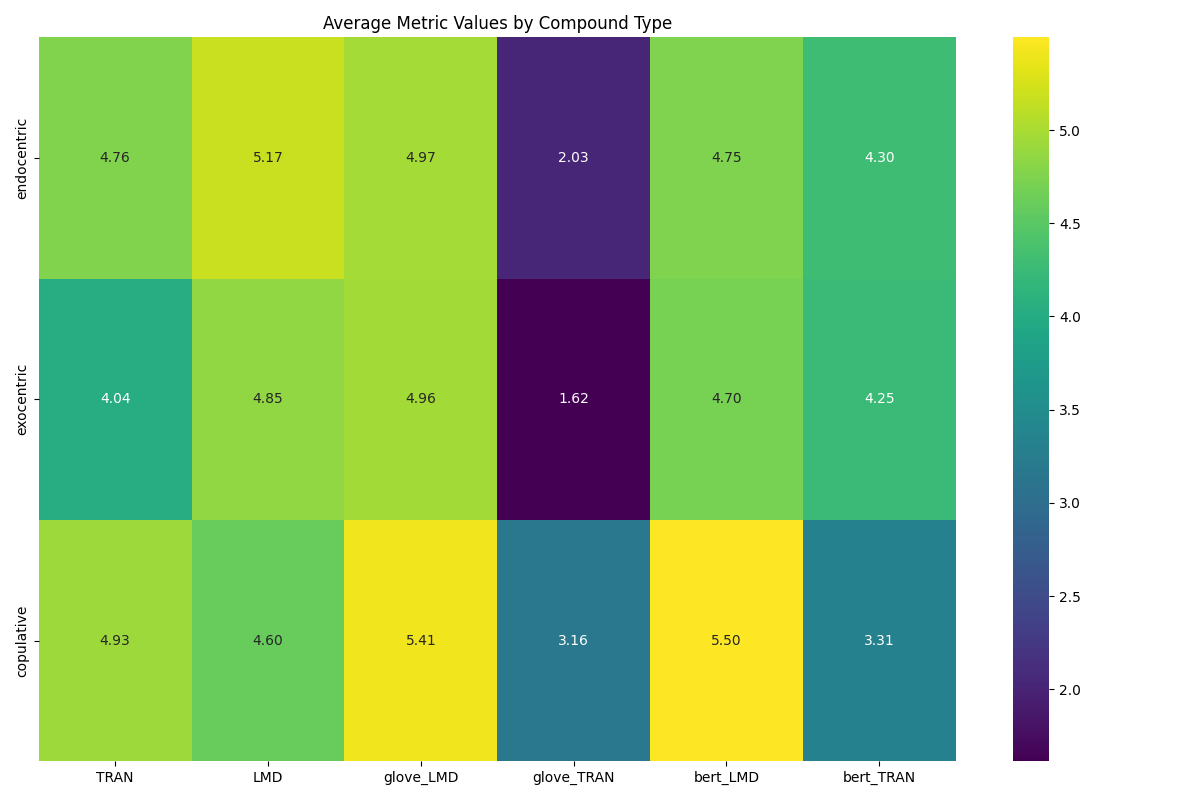}
    \caption{Compound Metrics Heatmap. TRAN refers to ST}
    \label{fig:cmetheat}
\end{figure*}

\section{Datasets}
Psycholinguistic dataset \cite{Juhasz2015} in processing containing 628 lexicalized English compounds annotated for LMD and ST.

We use Edinburgh Associative Thesaurus (EAT) \cite{ea-thesaurus} for word associations and LaDEC: Large database of English compounds\cite{Gagné2019} for predictability and BNC word frequency.

\begin{figure*}[t]
  \centering
  \includegraphics[width=0.65\textwidth]{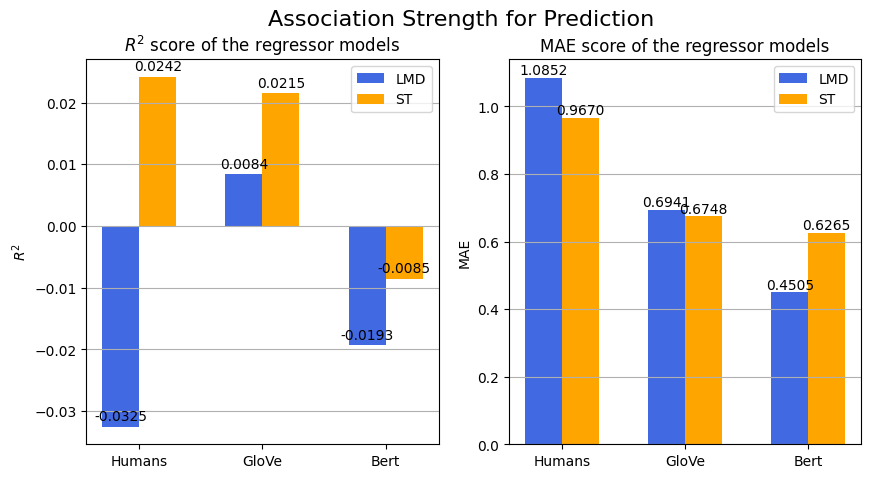}
  \includegraphics[width=0.65\textwidth]{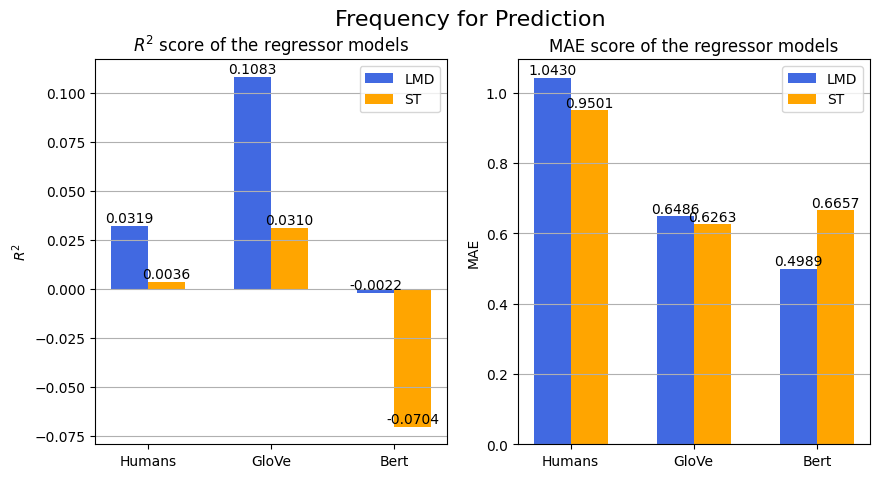}
  \includegraphics[width=0.65\textwidth]{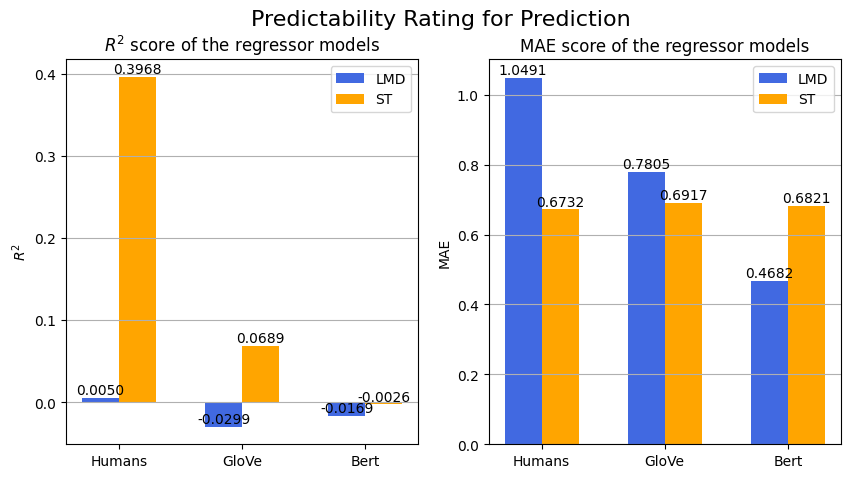}
  \caption{Performance of Regressors}
  \label{fig:regr}
\end{figure*}

\section{Results}
The MAE and Spearman correlation between the human judgments of LMD and ST and those derived from Glove and BERT embeddings matched the values mentioned in the main reference paper \cite{buijtelaar-pezzelle-2023-psycholinguistic}.

\subsection{Correlation}
\begin{table}
    \centering
    \begin{tabular}{c|lll}
    \hline
    \textbf{Factor} & \textbf{Humans} & \textbf{Glove} & \textbf{BERT}\\
    \hline
    {Association} & -0.0719 & -0.2415 & -0.0536 \\
    {Frequency} & -0.1395 & -0.0172 & 0.0636 \\
    {Frequency (R-L)} & \textbf{-0.1714} & \textbf{-0.4345} &\textbf{-0.2303 }\\
    {Frequency (R+L)} & -0.1585 &-0.0410 & 0.1023 \\
    {Predictability} & -0.1575 & -0.0657 & -0.0458\\
    \hline
    \end{tabular}
    \caption{Spearman correlation between \textbf{LMD} values and the factors}
    \label{tab:corr_lmd}
\end{table}
From the Table ~\ref{tab:corr_lmd} we can see that LMD had a negative correlation with all the factors. Among human-annotated values, predictability rating and frequency had a significant correlation. Only association are significantly correlated with Glove's values of LMD. 
In contrast, none of the linguistic factors we examined showed a significant correlation with the LMD values derived from BERT embeddings.
Frequency (R-L) had the strongest correlation across all the representations.

\begin{table}
    \centering
    \begin{tabular}{c|lll}
    \hline
    \textbf{Factor} & \textbf{Humans} & \textbf{Glove} & \textbf{BERT}\\
    \hline
    {Association} & 0.2365 & 0.2300 & 0.0281 \\
    {Frequency} & -0.0588 & \textbf{0.4410} & 0.2319 \\
    {Frequency (R-L)} & -0.0351 & 0.0091 & 0.0636 \\
    {Frequency (R+L)} & 0.0351 & -0.0306 & \textbf{0.2478} \\
    {Predictability} & \textbf{0.7326} & 0.3096 & 0.1033\\
    \hline
    \end{tabular}
    \caption{Spearman correlation between \textbf{ST} values and the factors}
    \label{tab:corr_st}
\end{table}
From the Table ~\ref{tab:corr_st} we can see that all the significant correlation between ST values and the factors are positive. Among human-annotated values, association strength is strongly correlated, followed by predictability strength. All three factors are significantly correlated with Glove's values of ST. Only the frequency and predictability ratings show a significant correlation with the ST values of the BERT embeddings.

\subsection{Regressors to Predict LMD and ST}
The graphs in Figure ~\ref{fig:regr} show the results of the regressors trained on the factors to predict the LMD and ST values. We can see that association strength is a poor predictor for both LMD and ST values. Frequency is only able to predict the LMD values from Glove embeddings. Predictability rating is a good predictor of only the ST values from human annotations.

\section{Discussion}

\subsection{Compound Type Distribution and Embedding Model Performance}

Our analysis reveals significant insights into both the distribution of compound types in English and how different embedding models capture their semantic properties. Figure \ref{fig:ctypedistro} shows the overwhelming predominance of endocentric compounds in our dataset (approximately 68\% endocentric vs. 31\% exocentric and <1\% copulative) confirms previous linguistic analyses of English compound formation preferences. Our dataset's composition, 68\% endocentric vs. 31\% exocentric—
is consistent with patterns observed in previous compound 
studies \cite{Libben1998}, though we note this 
reflects the sampling strategy of \citet{Juhasz2015} rather 
than a representative survey of English compounding. 
This distribution reflects English's tendency toward transparent, compositional word formation strategies, where the semantic head is explicitly represented within the compound.

\subsection{Semantic Transparency Across Compound Types}

The transparency (ST) metrics reveal patterns that largely align with theoretical predictions from morphological theory. Figure Endocentric compounds demonstrate higher transparency values (4.76) than exocentric compounds (4.04), confirming that head-modifier relationships contribute to semantic predictability. This finding supports \citet{Libben1998} transparency hierarchy and \citet{GagneSpalding2009} relational framework theories, which posit that compounds with clear internal semantic structures are more easily processed and interpreted.
The surprisingly high transparency value for copulative compounds (4.93) suggests that coordinate relationships may be particularly accessible to speakers, despite their relative rarity in English. This might indicate that the balanced semantic contribution from both constituents creates a unique form of transparency that differs from the asymmetrical relationship in endocentric compounds.

\subsection{Model-Specific Representations of Compound Semantics}

\subsubsection{Divergence Between BERT and GloVe}

The stark contrast between how BERT and GloVe represent compound transparency is one of our most striking findings. GloVe's transparency values are dramatically lower across all compound types (endocentric: 2.03; exocentric: 1.62; copulative: 3.16) compared to BERT's values, which more closely align with the original ST ratings. This suggests that contextual embeddings (BERT) may better capture the compositional nature of compounds than static embeddings (GloVe).
The divergence can be attributed to fundamental architectural differences: BERT's bidirectional, contextual nature allows it to better represent how compound meanings emerge from the interaction between constituents, while GloVe's context-independent vectors may struggle to capture these compositional semantics.

\subsubsection{Lexical-Morphological Distance Patterns}

The LMD metrics reveal a more complex picture than anticipated by straightforward compositional theories. Endocentric compounds show higher LMD values than expected (5.17), suggesting that even semantically transparent compounds maintain distinct representations from their constituents in embedding space. This supports dual-route theories of compound processing \cite{Kuperman2009}, which propose that compounds are accessed both as whole units and through individual units.

\section{Conclusion}
Our study confirms that contextualized embeddings (BERT) better mirror human semantic transparency judgments than static embeddings (GloVe), likely due to their capacity to model contextual interactions between morphemes. Predictability emerges as the most robust factor driving transparency, highlighting the role of semantic expectation in compound processing. These insights contribute to dual-route theories of morphological processing and inform the choice of embedding models for downstream applications.

\section{Limitations}
While our study sheds light on how static (GloVe) and contextualized (BERT) embeddings capture human semantic intuitions for English compounds, there remain several limitations:

\begin{itemize}
  \item \textbf{Language and Genre Coverage.}  
    We focus exclusively on lexicalized English compounds drawn from a psycholinguistic dataset of 628 items.  Our findings may not generalize to other languages (e.g., German, where compounding is more productive) or to less-frequent, novel compounds encountered in large-scale corpora.
  
  \item \textbf{Embedding Variants.}  
    Only one static embedding (GloVe) and one contextualized model (BERT\textsubscript{base}) were evaluated.  Future work should explore additional architectures (e.g., RoBERTa, ALBERT, or contextualized static hybrids) and compare multilingual or specialized domain embeddings.
  
  \item \textbf{Psycholinguistic Measures.}  
    We rely on pre-existing human ratings for lexeme meaning dominance (LMD) and semantic transparency (ST).  These measures come from a single study and may embed annotation biases or inter-rater variability that could influence our correlation and regression results.
  
  
  
  \item \textbf{Downstream Task Validation.}  
    Our evaluation metric is correlation with human judgments.  We do not assess the impact of compound representation quality on downstream tasks (e.g., machine translation, lexical semantic annotation), which is an important avenue for future validation.
\end{itemize}

\bibliography{custom}

\appendix

\section{Appendix}

\subsection{Graphs}

\begin{figure*}
    \centering
    \includegraphics[width=1\textwidth]{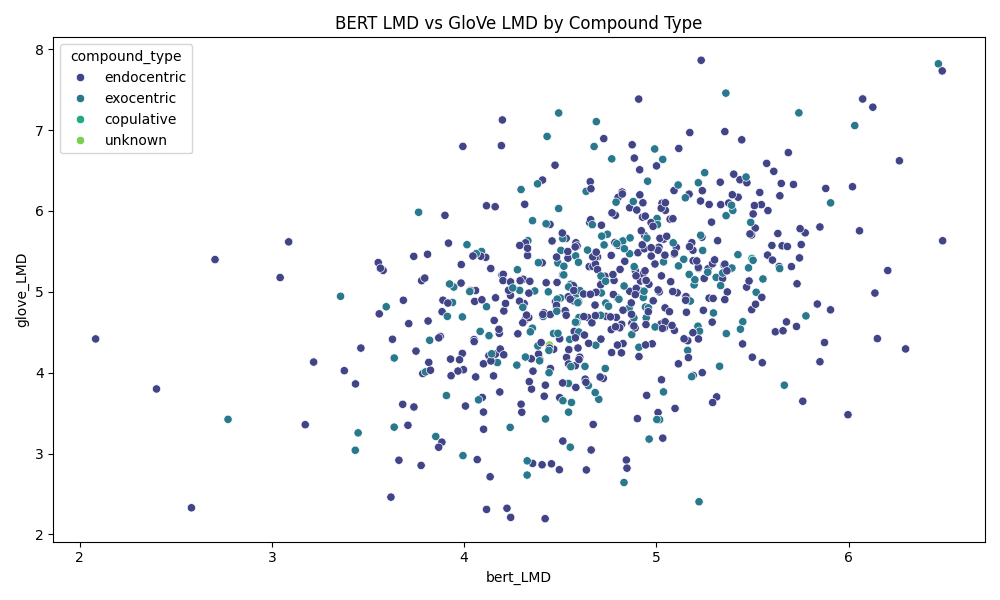}
    \caption{Bert vs GloVe LMD distribution}
    \label{fig:lmd}
\end{figure*}

\begin{figure*}
    \centering
    \includegraphics[width=1\textwidth]{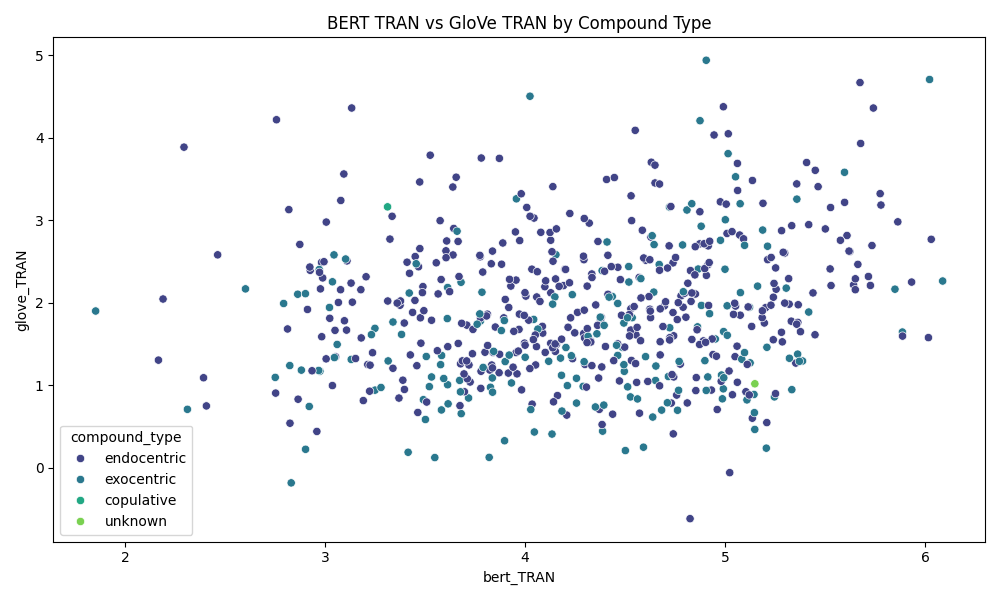}
    \caption{Bert vs GloVe LMD distribution}
    \label{fig:tran}
\end{figure*}



\end{document}